\documentclass[letterpaper, 10pt, conference]{ieeeconf}
\IEEEoverridecommandlockouts 

\usepackage{floatrow}
\usepackage{afterpage}
\usepackage{cite}
\usepackage{algorithm} 
\usepackage{algorithmic}  
\usepackage[algo2e]{algorithm2e} 
\usepackage{amsmath,amssymb,amsfonts}
\usepackage{algorithm}
\usepackage{graphicx}
\usepackage{textcomp}
\usepackage{booktabs}
\def\BibTeX{{\rm B\kern-.05em{\sc i\kern-.025em b}\kern-.08em
    T\kern-.1667em\lower.7ex\hbox{E}\kern-.125emX}}
    
    \usepackage{lineno}
\usepackage{lettrine}

\usepackage{url}
\usepackage{moreverb}

\usepackage{graphicx}
\usepackage{subcaption}
\usepackage{float}
\usepackage{amsmath}
\usepackage{array}
\usepackage{multirow}
\graphicspath{ {Figures/} }

\usepackage{dblfloatfix}
\usepackage[tableposition=top]{caption}

\usepackage{scalerel}
\usepackage{tikz}
\usetikzlibrary{svg.path}

\definecolor{orcidlogocol}{HTML}{A6CE39}
\tikzset{
  orcidlogo/.pic={
    \fill[orcidlogocol] svg{M256,128c0,70.7-57.3,128-128,128C57.3,256,0,198.7,0,128C0,57.3,57.3,0,128,0C198.7,0,256,57.3,256,128z};
    \fill[white] svg{M86.3,186.2H70.9V79.1h15.4v48.4V186.2z}
                 svg{M108.9,79.1h41.6c39.6,0,57,28.3,57,53.6c0,27.5-21.5,53.6-56.8,53.6h-41.8V79.1z M124.3,172.4h24.5c34.9,0,42.9-26.5,42.9-39.7c0-21.5-13.7-39.7-43.7-39.7h-23.7V172.4z}
                 svg{M88.7,56.8c0,5.5-4.5,10.1-10.1,10.1c-5.6,0-10.1-4.6-10.1-10.1c0-5.6,4.5-10.1,10.1-10.1C84.2,46.7,88.7,51.3,88.7,56.8z};
  }
}

\newcommand\orcidicon[1]{\href{https://orcid.org/#1}{\mbox{\scalerel*{
\begin{tikzpicture}[yscale=-1,transform shape]
\pic{orcidlogo};
\end{tikzpicture}
}{|}}}}

\usepackage[printonlyused]{acronym}

\usepackage[draft]{hyperref}
\usepackage{float}
\floatstyle{plaintop}
\restylefloat{table}
\begin{document}
\title{\LARGE \bf CD-TWINSAFE: A ROS-enabled Digital Twin for Scene Understanding and Safety Emerging V2I Technology}

\author{Amro Khaled$^{1,2\orcidicon{0009-0004-0617-6776}\,}$, Farah Khaled$^{1,2\orcidicon{0009-0001-6869-1288}\,}$,  Omar Riad$^{1,2\orcidicon{0009-0007-9128-4158}\,}$and Catherine~M.~Elias$^{1,2\orcidicon{0000-0002-1444-9816}\,}$,~\IEEEmembership{Member,~IEEE,}%
\thanks{*This work was not supported by any organization}
\thanks{$^{1}$C-DRiVeS Lab: Cognitive Driving Research in Vehicular Systems, Cairo, Egypt
{\tt\small cdrives.researchlab@gmail.com}}%
\thanks{$^{2}$Computer Science and Engineering Department - Faculty of Media Engineering and Technology - German University in Cairo, Egypt}%
\thanks{{\tt\scriptsize amrk13415@gmail.com,
farah.abdelghany@student.guc.edu.eg, riad.omar2004@gmail.com, catherine.elias@ieee.org}}%
}

\markboth{Journal of \LaTeX\ Class Files,~Vol.~14, No.~8, August~2015}%
{author1 \MakeLowercase{\textit{et al.}}:title here}
%



\maketitle
\begin{abstract}
In this paper, the CD-TWINSAFE is introduced, a V2I-based digital twin for Autonomous Vehicles. The proposed architecture is composed of two stacks running simultaneously, an on-board driving stack that includes a stereo camera for scene understanding, and a digital twin stack that runs an Unreal Engine 5 replica of the scene viewed by the camera as well as returning safety alerts to the cockpit. The on-board stack is implemented on the vehicle side including 2 main autonomous modules; localization and perception. The position and orientation of the ego vehicle are obtained using on-board sensors. Furthermore, the perception module is responsible for processing 20-fps images from stereo camera and understands the scene through two complementary pipelines. The pipeline are working on object detection and feature extraction including object velocity, yaw and the safety metrics time-to-collision and time-headway. The collected data form the driving stack are sent to the infrastructure side through the ROS-enabled architecture in the form of custom ROS2 messages and sent over UDP links that ride a 4G modem for V2I communication. The environment is monitored via the digital twin through the shared messages which update the information of the spawned ego vehicle and detected objects based on the real-time localization and perception data. Several tests with different driving scenarios to confirm the validity and real-time response of the proposed architecture.

\end{abstract}


\IEEEpeerreviewmaketitle
\section{Introduction}\label{sec1}
Autonomous vehicles (AV) have shown their effectiveness in reducing accidents, as, according to recent studies, human error contributes to 90\% of accidents \cite{abdel2024matched}. This highlighted the importance of advancements within the AV field, with a focus on safety as it was shown that, in addition to lower accidents risks, in the case of an accident the AV diminished the risk of injury from 98.71\% to 68.76\% when compared to human-driven vehicles\cite{abdel2024matched}. To achieve these privileges allowed by AVs, a perception system is paramount, to allow for the vehicle to be able to view and assess its surroundings. This was achieved through advances in camera-based perception systems, which allowed the AV to achieve object detection, depth estimation, and semantic segmentation. 

Despite these advancements in the AV field, an are that is under-represented is the use of the digital twin being integrated in a real-time replica of the AV and its operating environment. The use of digital twins in AVs is a rapidly evolving field, and although its initial uses were industry and aerospace related, they are recently being used more in the automation industry\cite{DENG2023180}. This is due to the importance in monitoring and controlling the AVs as well as providing a safety system for them.

In this paper, we introduce the CD-TWINSAFE,  Visual Perception-based V2I Digital Twin for Scene Understanding and Safety. CD-TWINSAFE takes a constantly running real-time stream from a stereo camera, then runs it through two perception models, which calculate several safety metrics such as the Time to Collision (TTC) and Time Headway (THW) and send them to a digital twin, which is responsible for sending a notification to the AV with a danger alert when certain critical thresholds are exceeded. 

The remainder of the paper is structured as follows. Section 2 reviews papers related to the topics covered in this paper. Section 3 covers the architecture of CD-TWINSAFE. Section 4 evaluates the approach in real-time scenarios, and section 5 discusses the conclusion of this paper, as well as future recommendations and limitations of this architecture.
\section{Related Studies}\label{sec1.5}

While many studies covered detection and depth estimation pipelines, rarely do these models rely on a purely visual system, while also achieve dependable accuracies in real-time systems. Qiao and Zulkernine\cite{masoumian2021absolute} used a stereo camera to concentrate on vehicle detection and geometric distance estimation. They compared the results of two detection models, namely You Only Look Once (YOLO)v4 and Faster-RCNN, and found that the using a real-time compatible system such as YOLOv4, which achieved 68 fps at a 99.16\% precision, compared to Faster-RCNN, which achieved 14 fps at 95.47\% precision, does not minimize the effectiveness of the model while performing at a real-time-compatible speed. They performed camera calibration and vanishing-point detection via Hough transforms, and then warped the image into bird's-eye perspective using a homography matrix, which enabled distance calculation up to 100 m. On the other hand, Masoumian et al.\cite{masoumian2021absolute} used deep learning with a monocular camera depth estimation, producing absolute distance prediction in real-time scenes. They also used YOLO for object bounding boxes, and then passed that bounding box to a quadratic mapping model that maps relative depth to real-world distances, and their pipeline accomplished a 96\% accuracy on a KITTI benchmark.

In recent years, A significant number of studies focused on digital twins and communication had an emphasis of using 2-way communication for the purpose of teleportation. Despite the varying hardware, vehicles in question and the choice of using an exiting simulator similarly to \cite{kim2024teleoperated} or tailoring one with the use of a game engine to fit their requirements as in \cite{wang2021digitalUnity,du2024tracked}, Robot Operating System (ROS) was either used for the communication framework or explicitly mentioned as a future work recommendation. On the other hand, few research papers have explored the usage of on-vehicle perception systems to convey scene understanding to a remote operator.



\section{Methodology}\label{sec2}

\begin{figure*}
    \centering
    \includegraphics[width=0.9\linewidth]{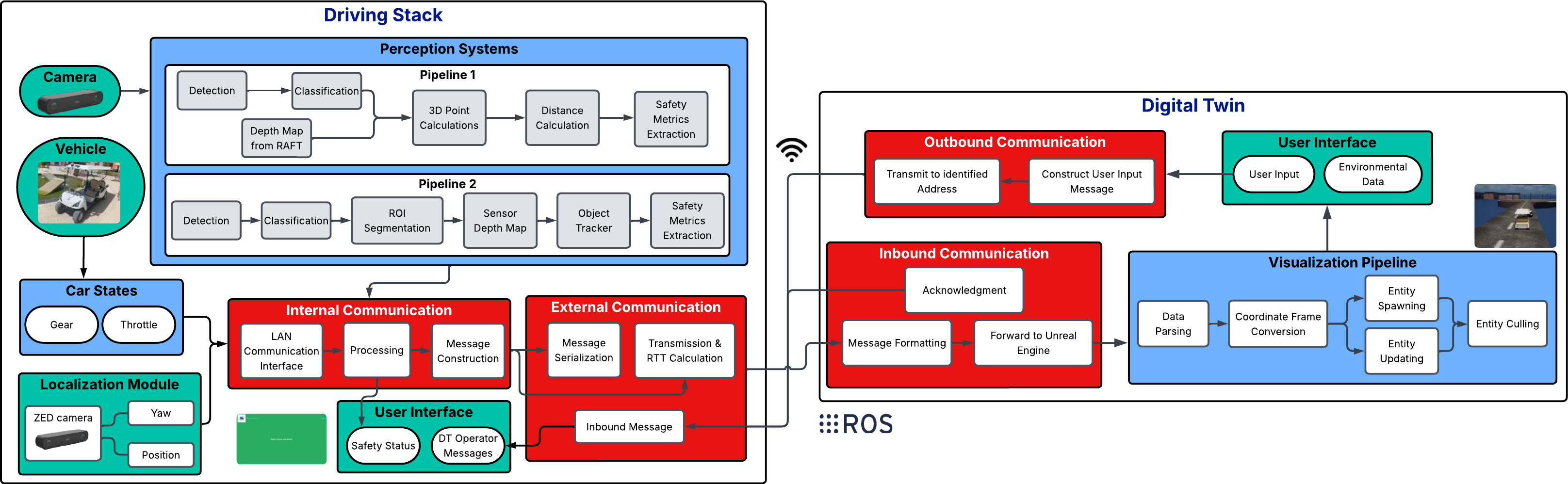}
    \caption{The CD-TWINSAFE Architecture Overview}
    \label{fig:big picture}
\end{figure*}

The developed architecture, CD-TWINSAFE, is composed of two main pipelines inspired from the architecture introduced by Elias et.al., G{\'o}mez-Hu{\'e}lamo et.al., and Manzour et.al. \cite{elias2022emerging,gomez2022build, elias2022cooperative,manzour2024development, guirguis2019ros}, as shown in figure  \ref{fig:big picture}, that communicate through UDP socket communication. The first is the perception and localization modules, and the second is the digital twin module responsible for visualizing the ego vehicle's perception and localization data.

\subsection{Perception and Localization Modules}
In this work, the vehicle is designed to be equipped with 2 main modules; localization and perception. Therefore, the vehicle is equipped with a number of sensors and processors that are used to perform the localization and perception modules. The added perception sensors are 2 Stereolabs ZED 2 camera with a resolution of $672 \times 376 px @ 10 FPS$ (neural depth mode) and $119.89 mm$ baseline. The 2 cameras are mounted on the vehicle roof on both sides with a spacing of $\pm 20 cm$ from the center of the vehicle and on a distance of $cm$ measured from the ground. The perception is done on 2 separate laptops which are used to run each studied pipeline solely. The specs of these laptops are: a laptop with Nvidia GeForce RTX 4060 GPU, and a laptop with NVIDIA GeForce RTX 3050 GPU.

\subsubsection{The Localization Module}
In this work, it is only necessary to measure the vehicle’s longitudinal velocity $V_{t}$ at each time sample $t$. This velocity is needed to compensate the relative measurement of the perception module. It is also important to mention that the time sample $t$ is determined based on the perception module rate. The vehicle velocity is obtained using direct measurement from the vehicle’ throttle with calibration using the gear state. This velocity measurement is essential in estimating the absolute velocity if the Preceding Vehicle $V_{{Abs}}$. 

\subsubsection{The Vision-based Perception Module}
For the implemented perception module, a layered pipeline composed of a number of sequential stages has been structured with four stages. Initially, the layered pipeline initializes the stereo camera, extracts, and stores its intrinsic parameters, such as the baseline, focal length, and principal points.

The first stage is the \textbf{Object Detection and Classification}. In this stage, all surrounding objects are detected, bounded in 2D boxes, and classified with one of the trained classes. The model is only trained to limited classes relevant to the proposed architecture, which are the "\textit{Car}" and "\textit{Pedestrian}" classes. Second, the \textbf{Depth Estimation} layer is activated which aims to estimate the depth of the detected objects relative to our ego vehicle.
Then comes the third stage; \textbf{Object Tracking \& Kinematic Estimation}, that is important to track the historical kinematical estimations of the object. This helps in extracting more kinematical properties as the object speed and yaw angle. Finally the forth stage which is \textbf{Safety Features Extraction} that is concerned with extracting and computing some important safety features which are Time-to-Collision (TTC) followed by Time-Headway (THW) and these features are the ones needed for the user interface.

\paragraph{Stage 1: Object Detection and Classification}
In the implementation of this stage, three different benchmark methods have been investigated. These methods are YOLOv8-n \cite{ultralytics2025yolov8}, Faster R-CNN \cite{ren2016fasterrcnnrealtimeobject}, and SSDLite \cite{liu2016ssd}. To quantitatively assess detector, the three methods have been examined on over 100 frames from a ZED SVO, logging per-frame latency, number of detections, and mean confidence. Table \ref{tab:detection_stats} summarizes the obtained results to prove the superiority of YOLOv8-n especially in terms of the latency which is critical in our full cooperative system scenario.

\begin{table}[H]
  \centering
  \caption{Detection Benchmark Summary (100-frame mean ± std).}
  \label{tab:detection_stats}
  \begin{tabular}{lccc}
    \toprule
    \textbf{Model}     & \textbf{Latency (ms)} & \textbf{\# boxes} & \textbf{Avg.\ conf.} \\
    \midrule
    YOLOv8-n          & $18.4 \pm 3.5$      & $2.2 \pm 0.4$      & $0.70 \pm 0.05$ \\
    Faster R-CNN       & $127.8 \pm 3.7$     & $33.1 \pm 6.7$     & $0.22 \pm 0.03$ \\
    SSDLite320         & $101.2 \pm 12.1$    & $300$ (fixed)      & $0.07 \pm 0.01$ \\
    \bottomrule
  \end{tabular}
\end{table}
At the end and based on the results, it is decided to continue with YOLOv8 in ByteTrack \cite{zhang2022bytetrackmultiobjecttrackingassociating} mode.

\paragraph{\textbf{\textit{Stage 2: Depth Estimation}} \label{sec:stage2} }
In the implemented methodology, the second and third stages, two different state-of-the-art pipelines have been investigated which are RAFT-Stero Depth Estimation and ROI-Based Segmentation and EMA Tracking

\textbf{Pipeline 1: RAFT-Stereo Depth Estimation}
The first investigated method is Recurrent All-Pairs Field Transforms (RAFT)-Stereo \cite{lipson2021raft}. RAFT works directly with the left and right images from the stereo camera. It takes both images as input and runs for 24 iterations to produce a high resolution disparity map $d$. This disparity map is comprised of point pairs $d(u,v)$, where each value corresponds to the shift between the left and right images. This disparity is then converted to depth map $Z$ in which each point depth $Z(u,v)$ is computed using the below formula:
\begin{equation}
Z(u,v) = \frac{fB}{d(u,v)}
\end{equation}
where $f$ is the focal length of the camera and $B$ is its baseline. 

This depth $Z(u,v)$ represents the distance between the camera frame and the point in the disparity map $d(u,v)$. The obtained point depth is further calibrated to compensate the $15^\circ$ tilt of the camera mounted on the vehicle. 

Moreover regarding the implementation of the forth stage -\textbf{Object Tracking \& Kinematic Estimation}-, an object tracker class is created to store the history of each object detected for a few frames, even after its disappearance from the camera view, to enhance the tracking of the objects and handle slight occlusions. Furthermore, the object's properties' history are saved including distance, speed, etc. 

At the end, for each detected object, its 2D coordinates are matched to the disparity map of the scene provided by RAFT-Stereo, in order to convert the coordinates to 3D to be later used for calculations. 

To conclude, few features are extracted for each object within the scene, such as the distance between the camera and object, the object's absolute speed in the scene with respect to the world frame, as well as the object's orientation relative to the camera frame. 

The distance of the object is considered the distance between its center and the camera, while its speed is calculated using the saved historical kinematical information as follows:
\begin{equation}
    v_{\mathrm{obj}}
  = - \frac{v_x\,X + v_z\,Z}{\sqrt{X^2 + Z^2}}
\end{equation}
where \(v_x\) and \(v_z\) is the change of the preceding vehicle's position in relation to the time in the $x$ and $z$ directions respectively, while the $X$ and $Z$ are its coordinates in the 3D camera frame. 

Regarding the preceding vehicle yaw estimation (the orientation), it is calculated by sampling the four centers of the bounding box, and then calculating the top, bottom, right and left center points of the bounding box. The horizontal and vertical depth differences of these points are then calculated and compared, where:
\begin{equation}
\Delta_{\text{vertical}} = d_{\text{bottom}} - d_{\text{top}},
\quad
\Delta_{\text{horizontal}} = d_{\text{right}} - d_{\text{left}}
\end{equation}

A discrete value is then assigned to the yaw of the preceding vehicle based on the difference between the vertical and horizontal depth differences using the set of equations:
\begin{equation}
\text{yaw}_{\text{box}} =
\begin{cases}
0^\circ, & \Delta_{\text{horiz}} < 0,\\
180^\circ, & \Delta_{\text{horiz}} > 0,
\end{cases}
\quad
\text{if }|\Delta_{\text{horiz}}| > |\Delta_{\text{vert}}|
\end{equation}

\begin{equation}
\text{yaw}_{\text{box}} =
\begin{cases}
90^\circ, & \Delta_{\text{vert}} < 0,\\
270^\circ, & \Delta_{\text{vert}} > 0,
\end{cases}
\quad
\text{if }|\Delta_{\text{vert}}|\ge|\Delta_{\text{horiz}}|
\end{equation}

The obtained yaw value is further used to determine if the preceding vehicle is parallel or perpendicular to  vehicle. The preceding vehicle is considered to be parallel, if the angle is \(0^\circ\) or \(180^\circ\), while if the obtained yaw angle is \(90^\circ\) or \(270^\circ\), then the preceding vehicle is considered to be perpendicular. 

\textbf{Pipeline 2: ROI-Based Segmentation and EMA Tracking}
For the second pipeline implementation, it is decided to work with segmented objects rather than just using bounding box as detection output. 

The estimator samples the center pixel of each bounding box from the full-frame 3D point cloud and converts millimeters to meters. Region-of-Interest (ROI) segmentation crops each detection window, applies a DeepLabV3-ResNet50 semantic segmentation model \cite{chen2017rethinking} to produce a binary mask for cars, and overlays the mask.  

The object tracker is implemented in an \texttt{ObjectTracker} class that tracks per-object depth and velocity with EMA smoothing, and also computes and smooths TTC and THW, handling infinities. For each track, it maintains a rolling window of recent depth measurements and computes a smoothed depth as the mean over that window. An anchor-based Vehicle is calculated across the window, then an exponential moving average with coefficient \(\alpha=0.3\) at 20 fps is applied. Raw TTC is calculated as the smoothed distance divided by the negative EMA velocity—set to \(\infty\) if the velocity is near zero or positive—and raw THW as the smoothed distance over ego speed—set to \(\infty\) if ego speed is below 0.1 m/s. Both TTC and THW raw values, computed in the following step, are EMA-smoothed when finite (retaining the previous EMA on \(\infty\)), then formatted as strings (e.g. “3.4s” or “inf”).  

\paragraph{\textbf{\textit{Stage 5: Safety Features Extraction}}}

After extracting the distance and speed, both pipelines implement the fifth and last layer to compute the needed safety features, namely TTC and THW.

\begin{equation}
  \mathrm{TTC} =
  \begin{cases}
    \dfrac{d}{\lvert v_{\mathrm{rel}}\rvert}, & v_{\mathrm{rel}}<0,\\[6pt]
    \infty, & v_{\mathrm{rel}}\ge0,
  \end{cases}
  \quad v_{\mathrm{rel}} = v_{\mathrm{ego}} - v_{\mathrm{obj}}.
\end{equation}

\begin{equation}
  \mathrm{THW} = \frac{d}{v_{\mathrm{ego}}}.
\end{equation}

These metrics—along with relative and absolute velocities—are then logged, rendered as per-object overlays and dashboard elements, and optionally broadcast over UDP.

\subsubsection{On-board User Interface}

The UI on board the vehicle is responsible for representing the safety of the surroundings of the vehicle to the driver, by displaying states (safe, hazardous, dangerous) according to the detected object's safety metrics (TTC, THW). The UI also includes the ability for the digital twin operator to send messages to the vehicle's driver to communicate the safety of the surroundings as well.

\begin{figure}[H]
    \centering
    \begin{subfigure}[t]{0.45\textwidth}
        \centering
        \includegraphics[width=\textwidth]{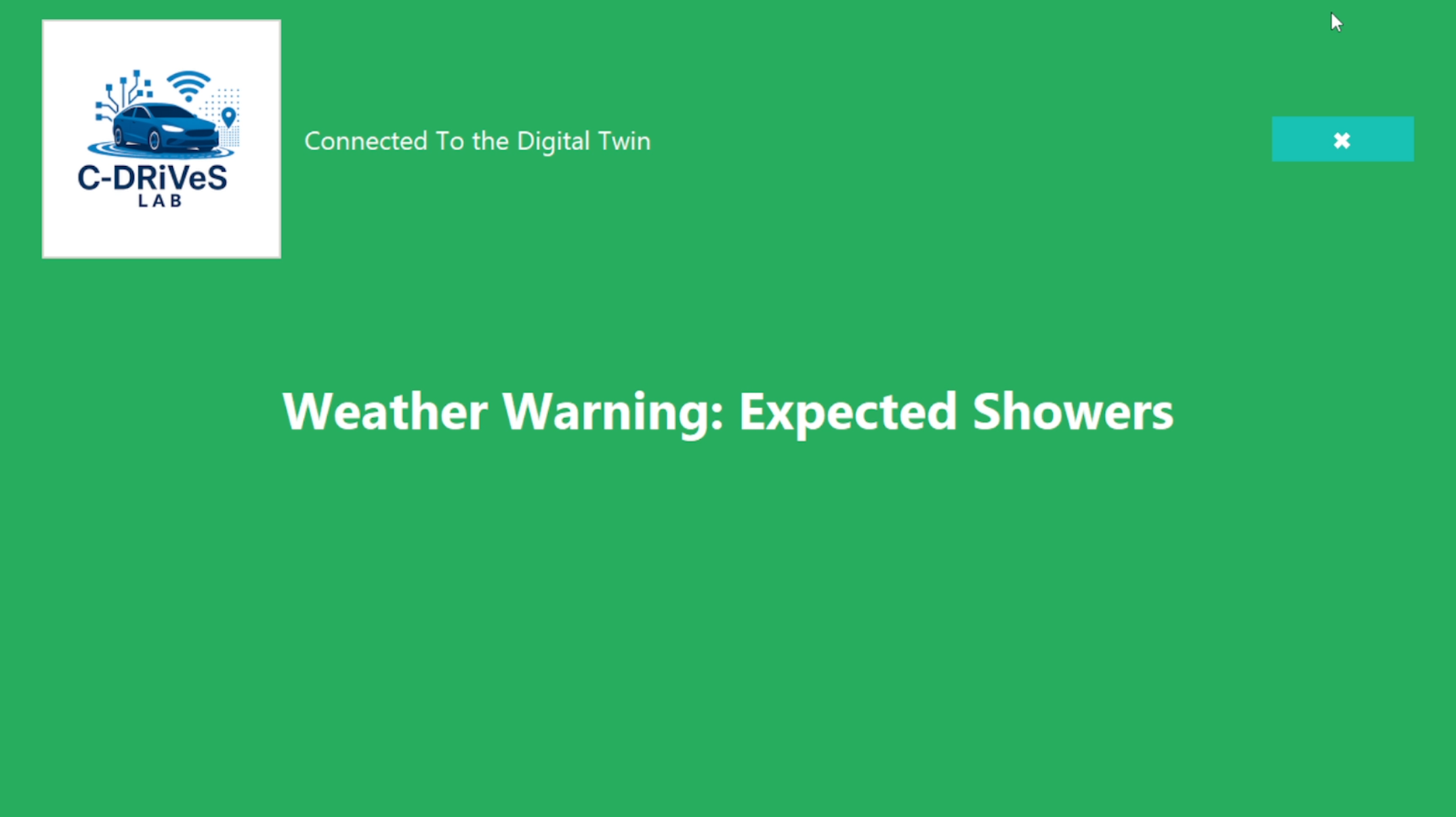}
        \caption{UI showing safe conditions with weather warning message.}
        \label{fig:subfig3}
    \end{subfigure}
    \hfill
    \begin{subfigure}[t]{0.45\textwidth}
        \centering
        \includegraphics[width=\textwidth]{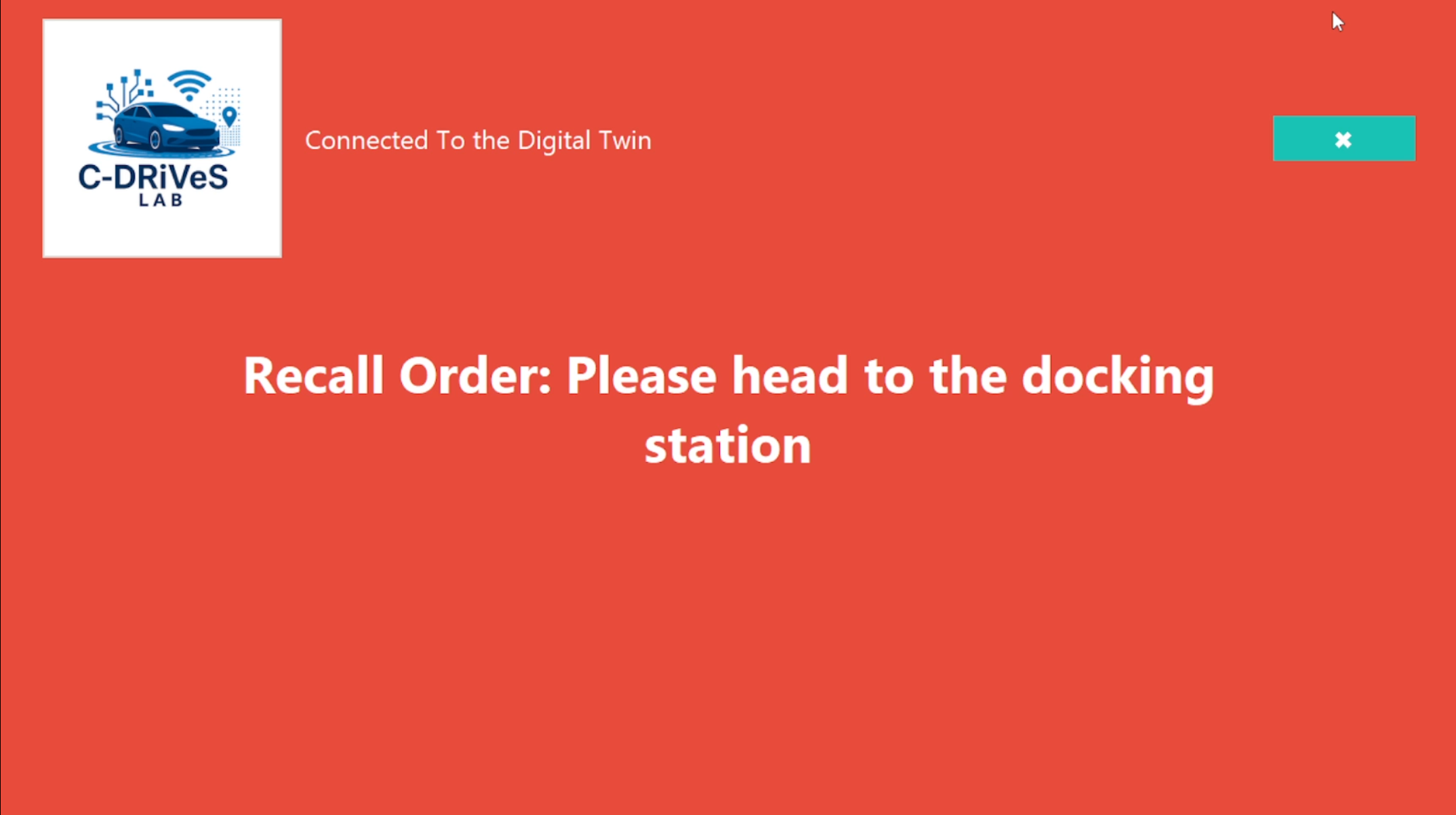}
        \caption{UI showing dangerous conditions with a recall order.}
        \label{fig:subfig4}
    \end{subfigure}
    \caption{On-board UI showing the state as background color and messages from digital twin operator.}
    \label{fig:result1}
\end{figure}

\subsection{Communication}

The communication from the driving stack to the digital twin is comprised of two main parts, internal and external communication. The internal communication module takes as input the the localization data and the perception data. 

Cellular communication technology, namely 4G, was used for the vehicle communication endpoint for its wide-area coverage as well as operational latency that is within the boundaries of real-time communication as shown in \cite{kamtam2024network}.

\subsubsection{On-Board Communication}

Communication between different modules within the ego vehicle is either done through the use of ROS2 nodes purely or UDP sockets for LAN communication integrating into the ROS2 network through a node based on the capabilities of the module in question.

\subsubsection{V2I Protocols}

The 2-way communication between the ego vehicle and the digital twin is established using standard UDP Sockets over the internet when the ego vehicle's on-board 4G device contacts the digital twin server; this is because 4G devices are typically connected to the internet through Carrier-Grade Network Address Translation (CG-NAT), which prevents them from having a publicly accessible IP address until they initiate outbound communication.

Furthermore, a method of using the ROS2 Data Distribution Service (DDS) serialization middleware is used in collaboration with custom ROS2 messages in order to create a scalable yet efficient message for over-the-network communication. Serialization occurs within a ROS node on the ego vehicle whereas the de-serialization is done on the receiving end of the communication through another ROS node, given that the message format is shared prior to communication as illustrated in Fig. \ref{fig:message}.

\subsection{Infrastructure - Digital Twin}

Unreal Engine (UE) 5  was chosen to create a digital twin capable of communication with the ego vehicle as well visually representing its perception and localization data in real-time.

\subsubsection{Class Structure}

A class structure was created for supporting the required functionality following the concepts of Object Oriented Programming (OOP).

\begin{figure*}[t]
    \centering
    \begin{subfigure}[b]{0.45\textwidth}
    \centering
    \includegraphics[width=1\linewidth]{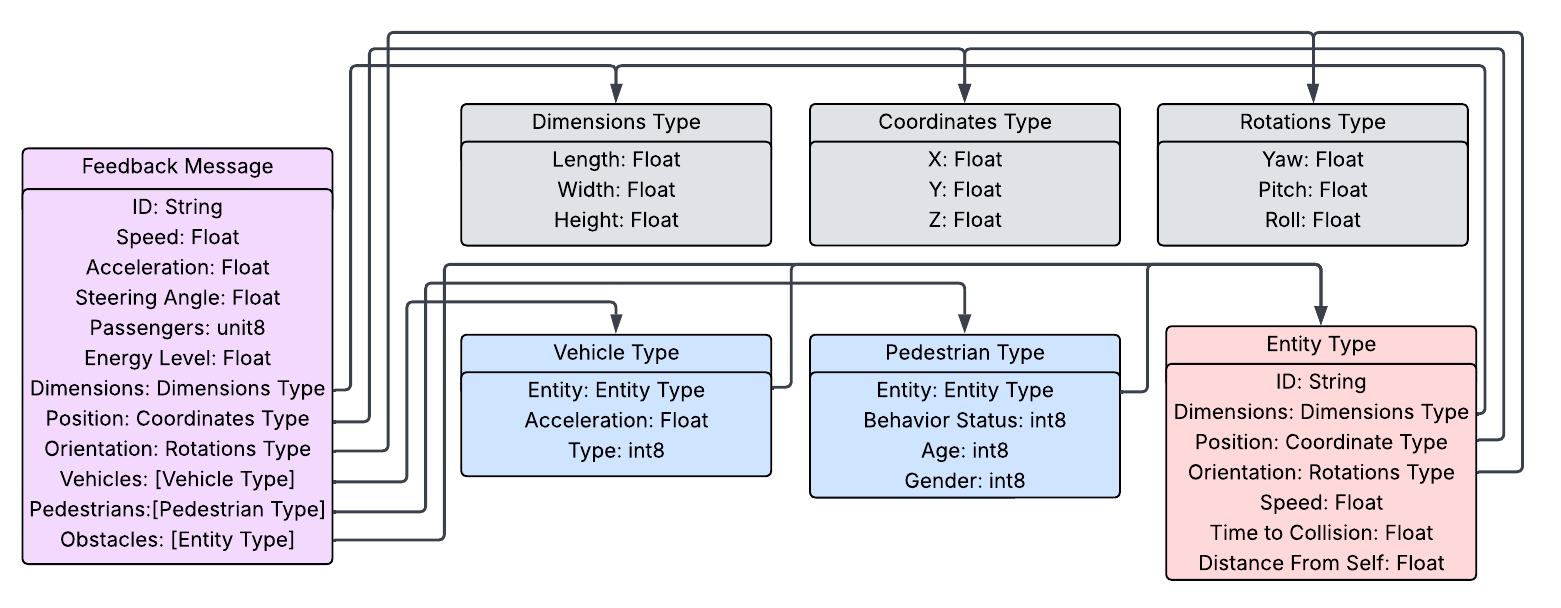}
    \caption{}
    \label{fig:message}
    \end{subfigure}
    \begin{subfigure}[b]{0.45\textwidth}
    \centering
    \includegraphics[width=1\linewidth]{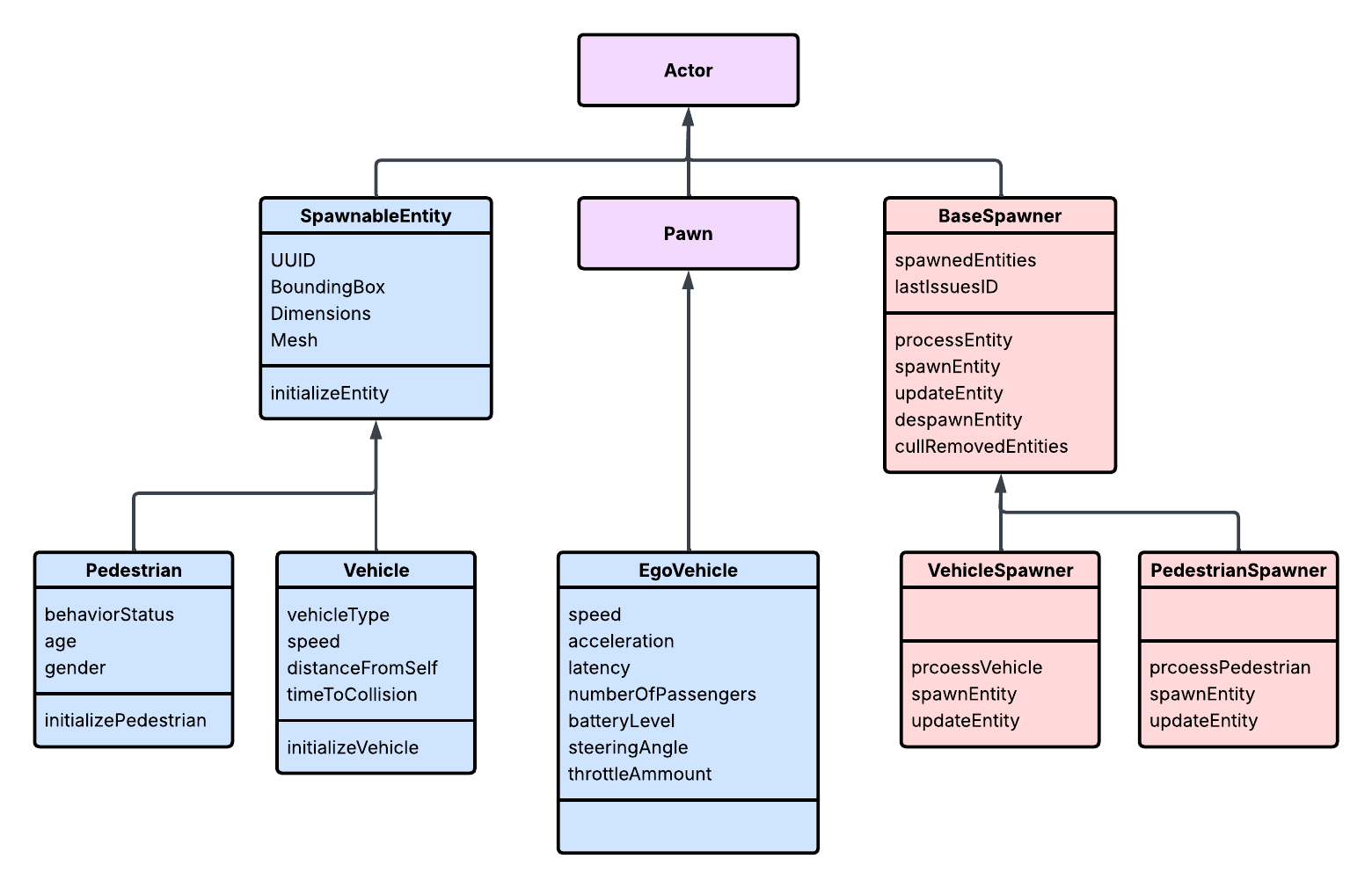}
    \caption{}
    \label{fig:class struct}
    \end{subfigure}
    \caption{(a) Message Format Used for Byte Encoding, (b) Implemented Class Structure of Digital Twin}
    \label{fig:combined}
\end{figure*}

The actor class of UE was used in order to place objects containing only logic (Spawners), visual aspects, or both. The pawn class was used to allow for the Ego-vehicle class to interface with the user input as well as provide the primary point of view (POV) for the operator as shown in Fig. \ref{fig:class struct}

\subsubsection{Visualization Pipeline}

The perception and localization data received goes through a few stages of processing in order to be properly visualized withing UE's 3D environment starting with parsing the data received into containers stored in memory for further processing. 

Following that, the vehicle's latitude and longitude localization data is converted to the corresponding coordinates within the UE coordinate frame; furthermore, the perception data is converted from the ego vehicle's relative frame of reference to UE's coordinate frame with the transformation matrix:\\

\resizebox{\columnwidth}{!}{%
$
R_{zxy}(\phi, \theta, \psi) =
\begin{bmatrix}
\cos\psi \cos\theta & \cos\psi \sin\theta \sin\phi - \sin\psi \cos\phi & \cos\psi \sin\theta \cos\phi + \sin\psi \sin\phi \\
\sin\psi \cos\theta & \sin\psi \sin\theta \sin\phi + \cos\psi \cos\phi & \sin\psi \sin\theta \cos\phi - \cos\psi \sin\phi \\
-\sin\theta         & \cos\theta \sin\phi                              & \cos\theta \cos\phi
\end{bmatrix}
$
}

\begin{equation}
\label{eq:offset eq}
\overrightarrow{\text{POS}}_{\text{Object}} =
\overrightarrow{\text{POS}}_{\text{Ego Vehicle}} -
R_{zxy}(\phi, \theta, \psi)
{\begin{bmatrix}
x_{\text{Relative}} \\
y_{\text{Relative}} \\
z_{\text{Relative}}
\end{bmatrix}}^T
\end{equation}

Following the coordinate conversion, each entity is sent to its corresponding spawner; if it isn't an object that was previously being maintained, meaning it has a new detection ID, a new entity is created and supplemented with the object data and the appropriate mesh. On the other hand, if an object's ID exists within the spawner, its existing entity attributes are updated based on the received data. Finally, any objects not spawned or updated, but exist in the scene are due for removal by their spawner.

\subsubsection{User Interface}

The digital twin's user interface (UI) is straightforwards and has two main functionalities, the first is displaying detected objects accurately in 3D space as well as the ego vehicle's state data. The second is communicating an operator's text of hotkey input to the ego vehicle's user interface to inform the driver of road conditions or upcoming safety hazards, such as weather warnings or road status updates.

\begin{figure}[H]
    \centering
    \includegraphics[width=0.9\linewidth]{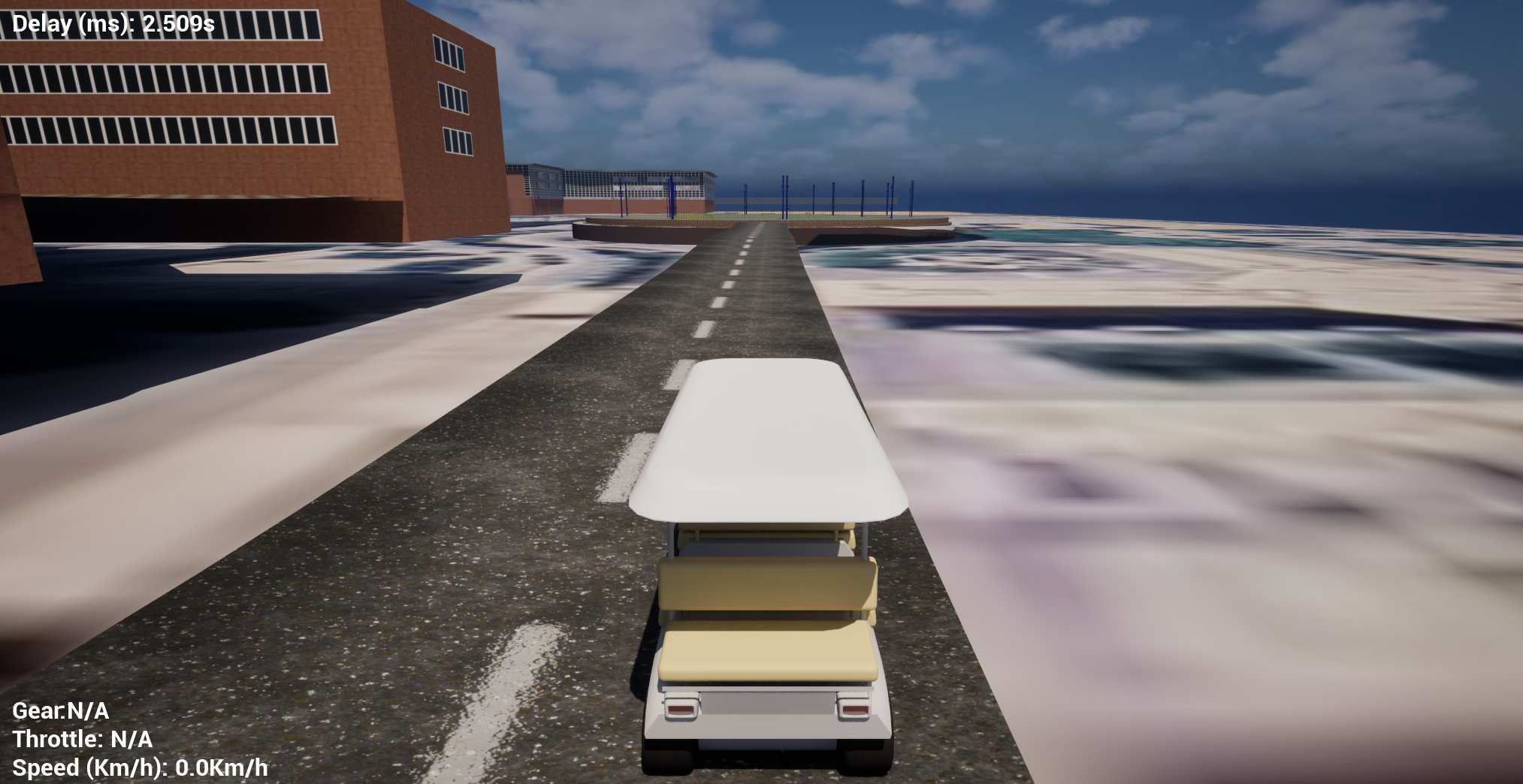}
    \caption{Snapshot of Digital Twin UI}
    \label{fig:dt ui}
\end{figure}

\begin{table*}[t]
\centering
\caption{Latency Statistics of the Proposed Communication Setup.}
\label{tab:latency_comp}
\begin{tabular}{cccccccccc}

\hline
\textbf{\begin{tabular}[c]{@{}c@{}}Transmitter\\ System\end{tabular}} & \textbf{\begin{tabular}[c]{@{}c@{}}Transmission\\ Technology\end{tabular}} & \textbf{\begin{tabular}[c]{@{}c@{}}Receiver\\ Syetem\end{tabular}} & \textbf{\begin{tabular}[c]{@{}c@{}}Receiver\\ Technology\end{tabular}}  & \textbf{\begin{tabular}[c]{@{}c@{}}Latency\\ Minimum\end{tabular}} & \textbf{\begin{tabular}[c]{@{}c@{}}Latency\\ Maximum\end{tabular}} & \textbf{\begin{tabular}[c]{@{}c@{}}Latency\\ Mean\end{tabular}} & \textbf{\begin{tabular}[c]{@{}c@{}}Standard \\Deviation\end{tabular}} & \textbf{\begin{tabular}[c]{@{}c@{}}Message\\ Loss Rate\end{tabular}} \\
\hline
WSL2 & 4G & WSL2 & Fiber Optics & 21.946 & 186.631 & 38.213 & 28.177 & 2.795\% \\
\hline

\end{tabular}%
\vspace{0.5em}
\end{table*}

\section{Results}\label{sec4}
\begin{figure}[H]
    \centering
    \begin{subfigure}[b]{0.45\textwidth}
        \centering
        \includegraphics[width=\textwidth]{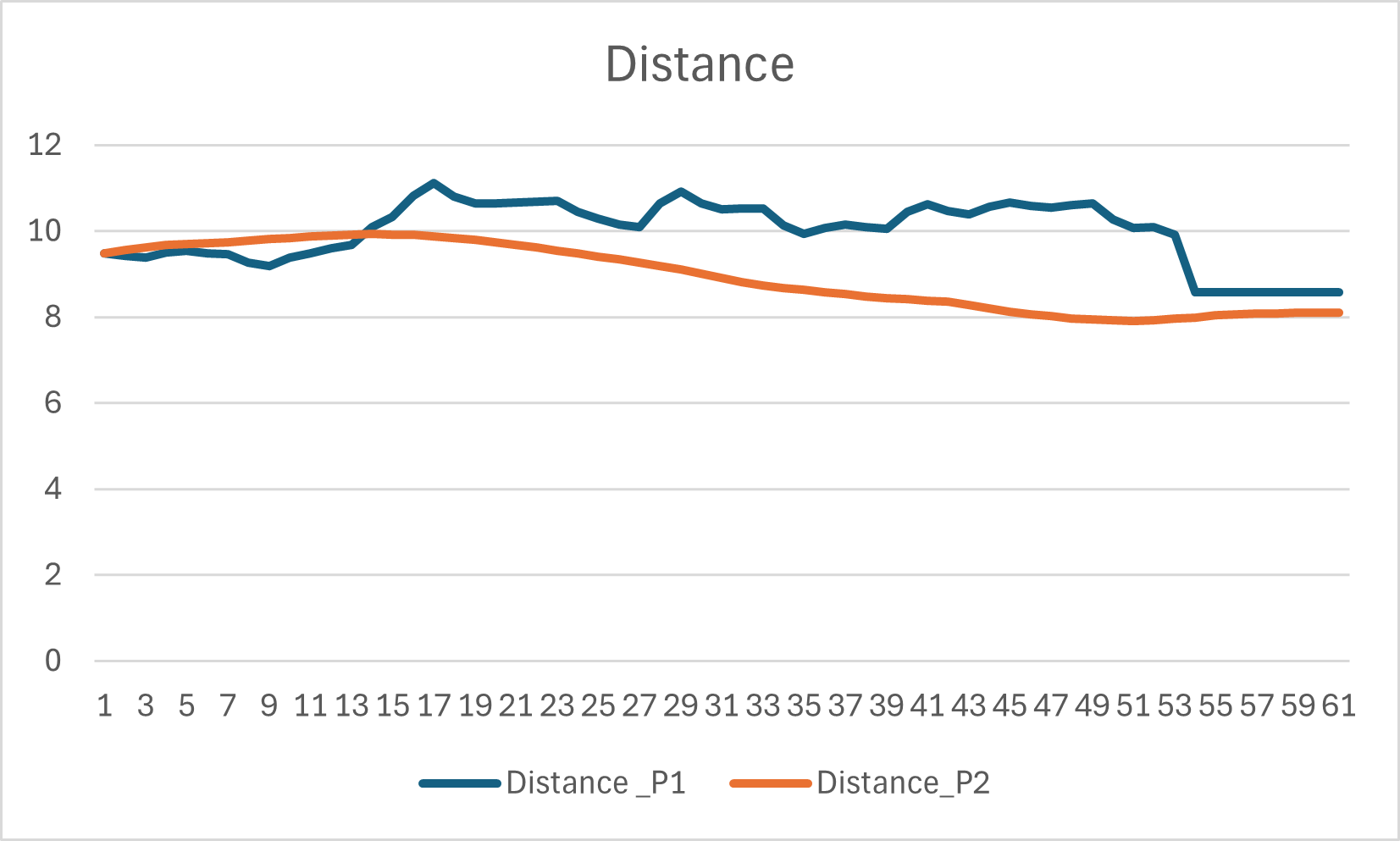}
        \caption{Distance values}
        \label{fig5:subfig1}
    \end{subfigure}
    \hfill
    \begin{subfigure}[b]{0.45\textwidth}
        \centering
        \includegraphics[width=\textwidth]{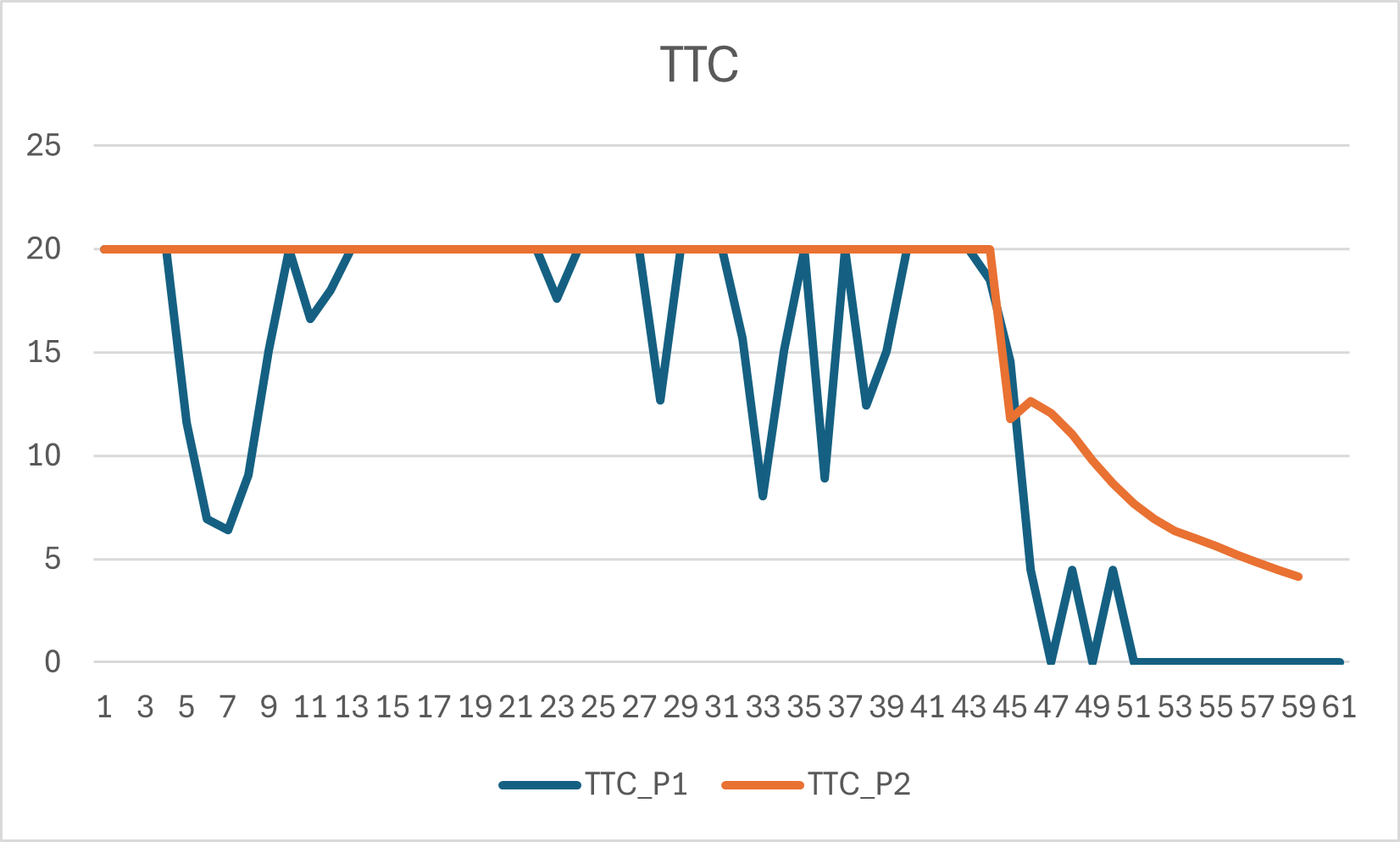}
        \caption{TTC values}
        \label{fig5:subfig2}
        \end{subfigure}
           \caption{Values of perception pipeline 1 and 2 in comparison}
    \label{fig:ValuesResult}
    \end{figure}

When comparing the results of the perception pipelines 1 and 2, the results are shown in figure \ref{fig:ValuesResult}.
\vspace{-15pt}
As shown in the graphs, the results, while close in values, show that Pipeline 2 was more accurate at larger distances, as shown in \ref{fig:ValuesResult}, while Pipeline 1 showed less stuttering and a smoother result in both \ref{fig5:subfig1} and \ref{fig5:subfig2}.

Furthermore, on the communications front, the proposed communication protocol was recorded to have very low network latencies, albeit with some jitter and loss. Such low-latency communication when combined with the protocol's low-bandwidth message format can aid in transmitting scene understanding data to the remote operator under sub-optimal network conditions as in Table \ref{tab:latency_comp}.
 
\begin{figure}[H]
    \centering
    \begin{subfigure}[b]{0.45\textwidth}
        \centering
        \includegraphics[width=\textwidth]{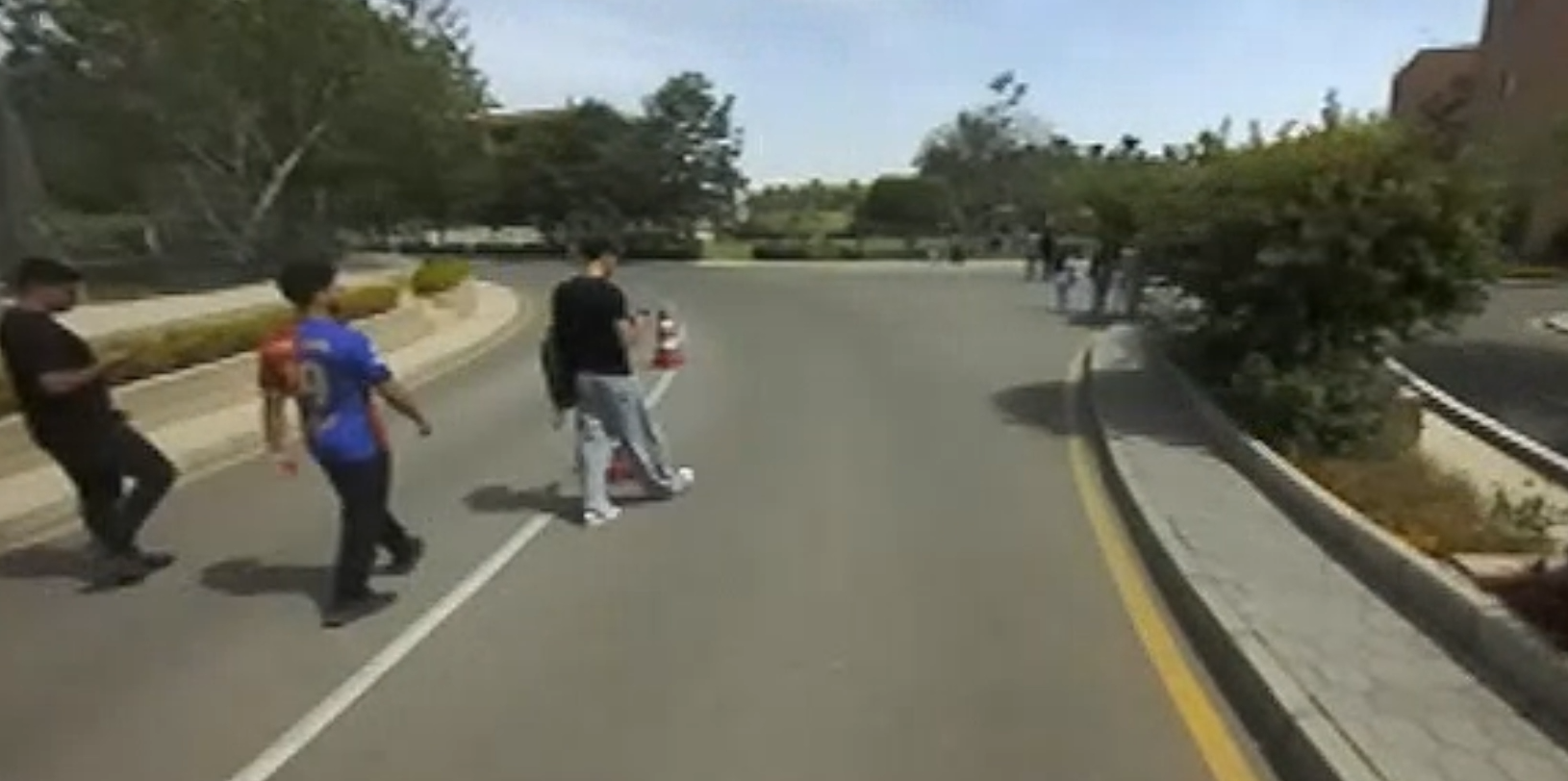}
        \caption{View from the camera}
        \label{fig:subfig2.1}
    \end{subfigure}
    \hfill
    \begin{subfigure}[b]{0.45\textwidth}
        \centering
        \includegraphics[width=\textwidth]{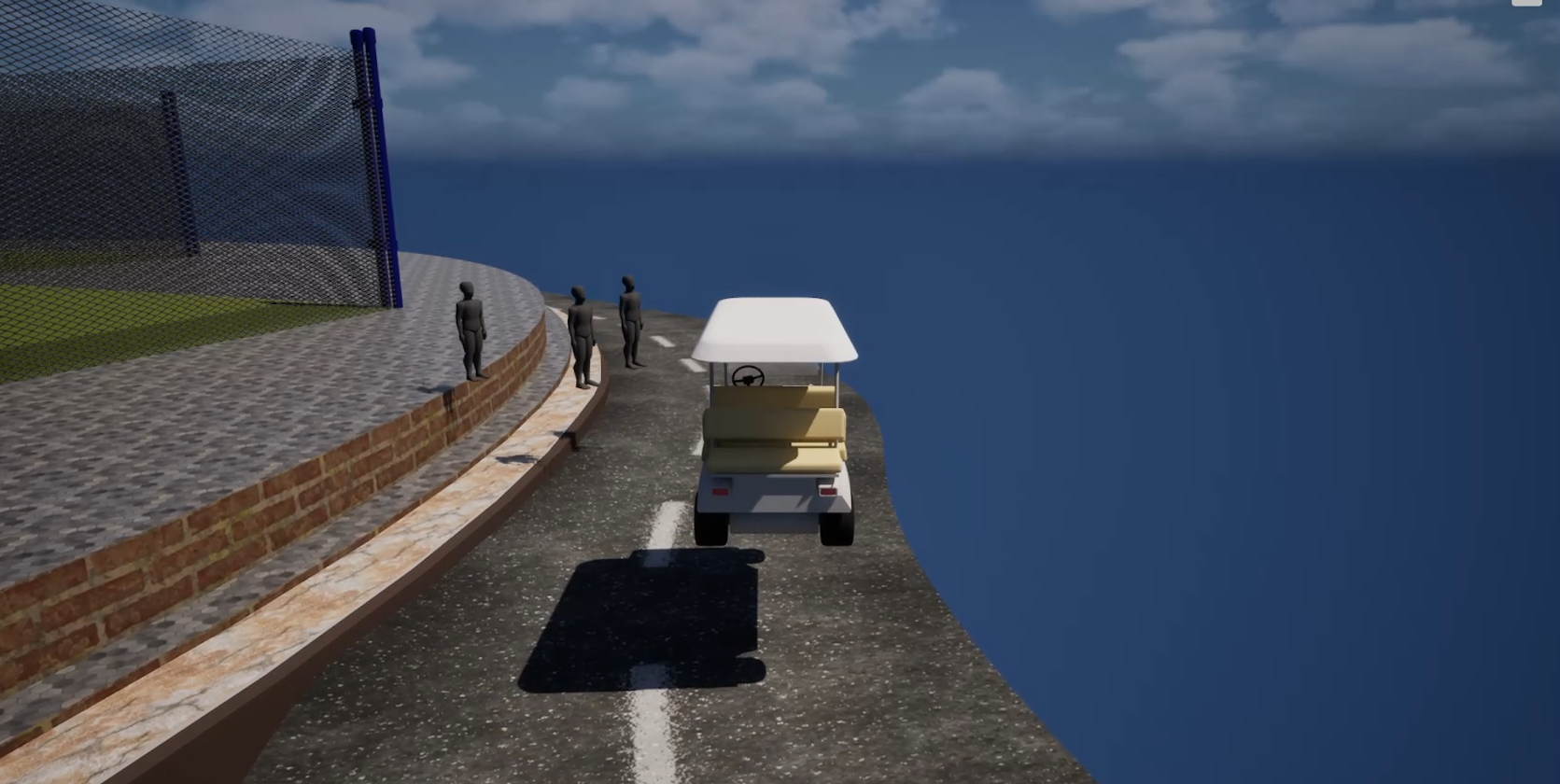}
        \caption{View from the digital twin}
        \label{fig:subfig2.2}
    \end{subfigure}
    \caption{Digital Twin displaying pedestrians.}
    \label{fig:result2}
\end{figure}

The full CD-TWINSAFE architecture was initially tested with pedestrians in a testing environment,  due to the importance of demonstrating all surrounding relevant objects to the DT operator. The results of these initial experiments are shown in figure \ref{fig:result2} and the experiment is further demonstrated through the video: \nolinkurl{https://youtu.be/CH68vn2g0B4}. However, the focus of further experimentation was more towards vehicles, as they posed more of a risk when considering vehicles on the road.
\vspace{-10pt}
\vspace{-15pt}
\begin{figure}[H]
    \centering
    \begin{subfigure}[t]{0.45\textwidth}
        \centering
        \includegraphics[width=\textwidth]{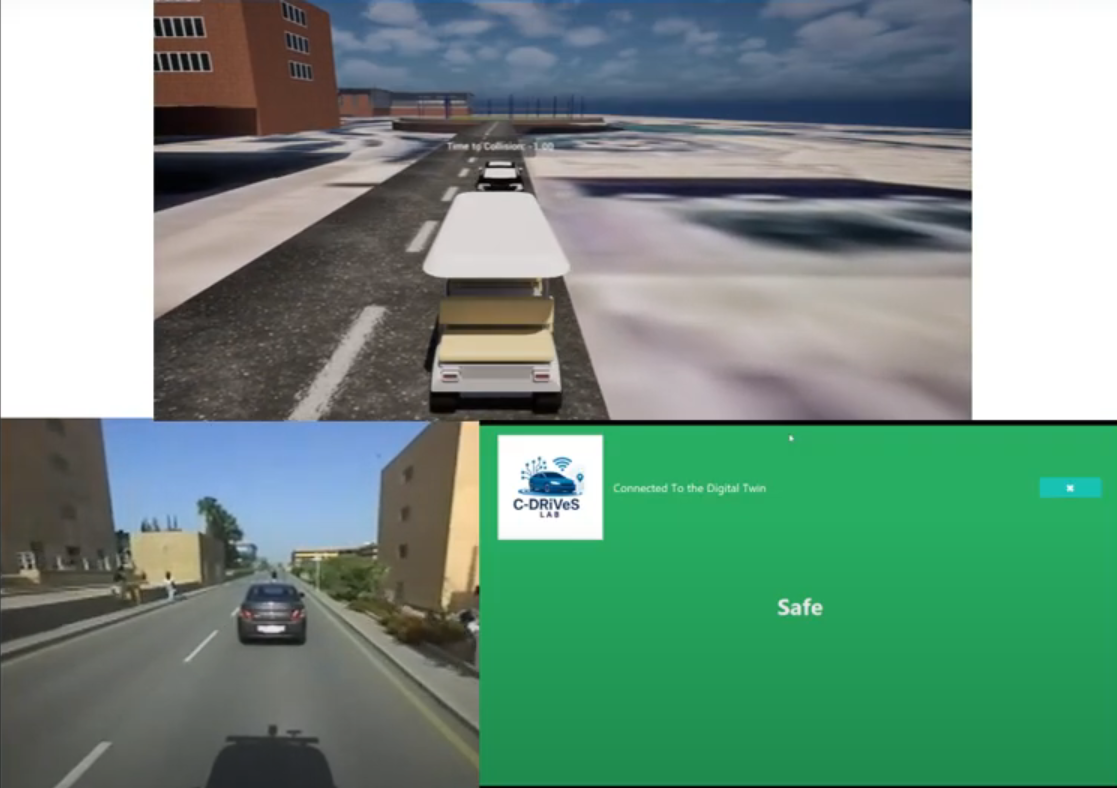}
        \caption{Safe conditions, with vehicle detected}
        \label{fig:subfig1}
    \end{subfigure}
    \hfill
    \begin{subfigure}[t]{0.45\textwidth}
        \centering
        \includegraphics[width=\textwidth]{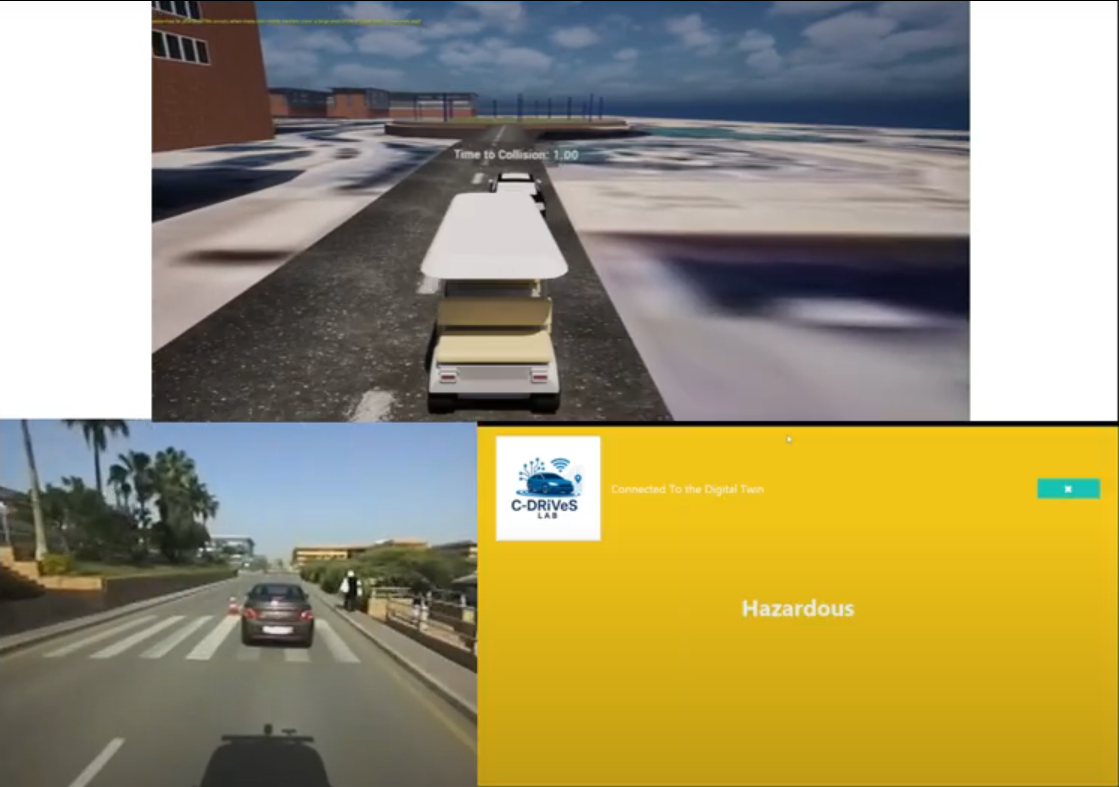}
        \caption{Ego vehicle is close to colliding, warning system alerts}
        \label{fig:subfig2}
    \end{subfigure}
    \hfill
    \begin{subfigure}[t]{0.45\textwidth}
        \centering
        \includegraphics[width=\textwidth]{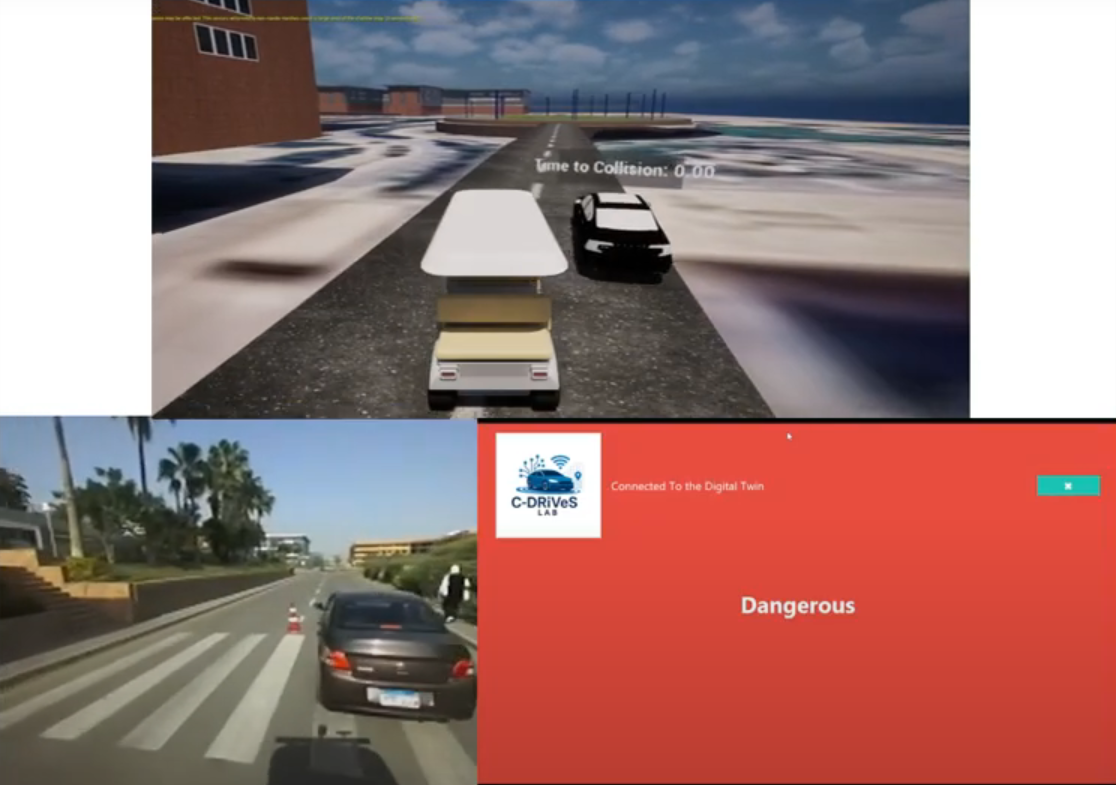}
        \caption{Dangerous TTC between ego vehicle and car detected, warning system alerts}
        \label{fig:subfig3}
    \end{subfigure}
    \hfill
    \begin{subfigure}[t]{0.45\textwidth}
        \centering
        \includegraphics[width=\textwidth]{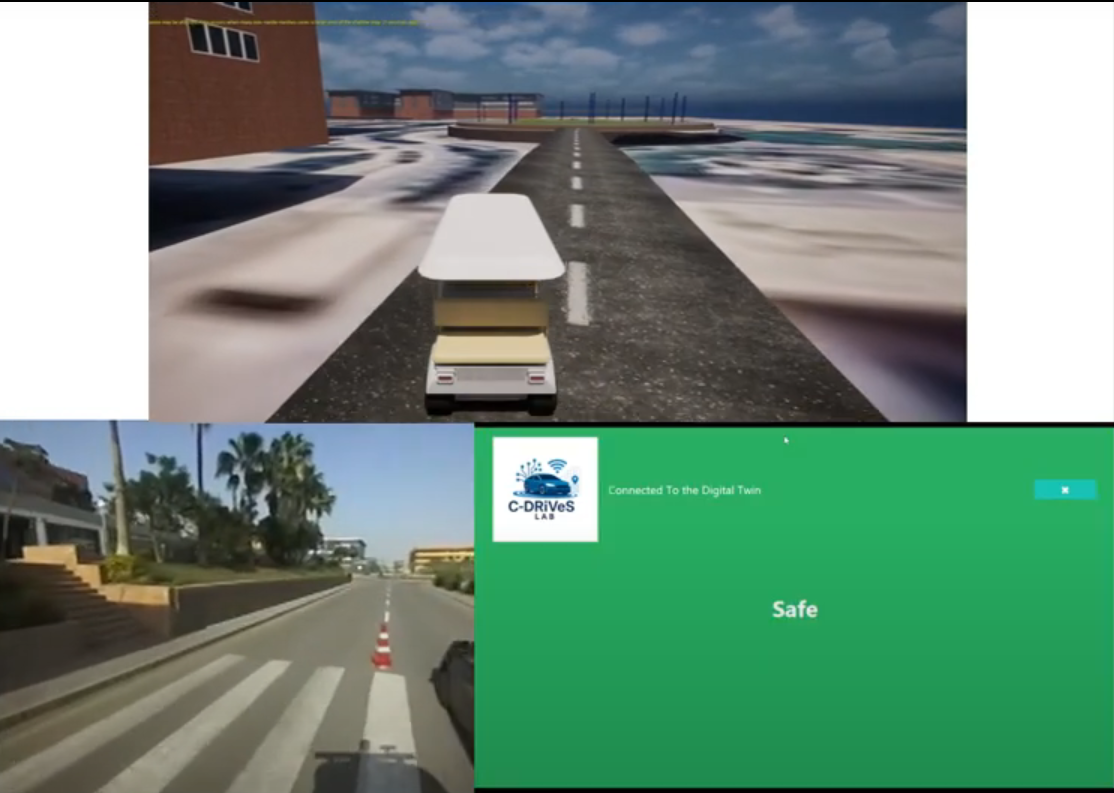}
        \caption{Avoided collision, warning system is back to safe}
        \label{fig:subfig4}
    \end{subfigure}
    \caption{Driving stack alert system responding to detections, as well as digital twin representation.}
    \label{fig:result1}
\end{figure}
\vspace{-15pt}
To further expand on the use of the pipeline in providing an understanding of the surrounding scene, an experiment was conducted, as shown in figure \ref{fig:result1}. The premise of the experiment was to provide the ego vehicle with a preceding vehicle, and test all three possibilities, a safe, hazardous and dangerous situation based on the set parameters. An on-board UI alert system was also available to display the values, when calculated by the perception system and classified by the digital twin, to provide further insight to passengers on board the vehicle. The experiment goes as follows: initially, the vehicle preceding is at a safe distance and speed relative to the ego vehicle, which translates to safe for the alert system. The speed of the preceding vehicle decreases gradually, thus decreasing the relative distance and becoming hazardous then dangerous, as shown in \ref{fig:subfig2} and  \ref{fig:subfig3} respectively.  The ego vehicle then avoids collision, and the scene becomes vacant, which prompts the alert system to display safe surroundings. Further demonstration of the experiment is seen through the video: \nolinkurl{https://youtu.be/puXVz62_36U}.

\vspace{-4pt}
\section{Conclusion and Future Recommendations}\label{sec4}
In conclusion, this paper aims to cover the use of perception systems paired with the digital twin model for achieving safety conditions of an autonomous vehicle. We achieved this by creating an on-board UI that takes as an input the perception parameters to assess the safety of the vehicle surroundings, as well as takes messages from an overlooking digital twin operator to alert the vehicle driver of weather and road conditions. The performance of CD-TWINSAFE is constrained by three key factors: first, our GPS receiver provides fixes at only 1 Hz, and the ZED 2's onboard IMU suffers from bias drift, leading to pose lag and accumulated error over extended operation. Second, the system's exclusive reliance on passive stereo vision renders disparity estimation and object detection vulnerable to challenging lighting—such as low light, direct glare, or deep shadows—which can produce false negatives or unstable depth maps. Third, the ZED 2 camera’s reliable depth measurements extend only to around 15 m, beyond which disparity becomes too small and noisy for accurate long-range tracking.

For future recommendations to  enhance the CD-TWINSAFE, it is recommended to expand the diving stack to achieve a control-capable vehicle to allow the digital twin operator to maneuver the vehicle from difficult situations remotely. Another recommendation is to fuse other sensors, such as the lidar and the radar, to reach a more robust system that veers from pure vision system limitations.
\appendices

\bibliographystyle{IEEEtran}
\bibliography{ref/ref.bib} 

\end{document}